\documentclass[conference,letterpaper]{IEEEtran}

\usepackage[utf8]{inputenc} 
\usepackage[T1]{fontenc}
\usepackage{url}
\usepackage{ifthen,enumitem}
\usepackage{cite}
\usepackage[cmex10]{amsmath} 
\usepackage{amsmath,amssymb,amsfonts,subcaption}
\usepackage{bbm,mathtools,xcolor}

\renewcommand{\epsilon}{\varepsilon}
\newcommand{\predspace}{\mathcal{Y}}
\newcommand{\lambdawcrc}{\lambda^{\!\rm W\text{-}CRC}}
\newcommand{\Gammawcrc}{\Gamma_{\theta}^{\rm W\text{-}CRC}}
\newcommand{\Lambdawcrc}{\Lambda^{\!\rm W\text{-}CRC}(\Dtr)}

\newcommand{\Dfull}{\mathcal D}
\newcommand{\lambdaythp}{\lambda_{y}^{\theta'}}
\newcommand{\lambdayth}{\lambda_{y}^{\theta}}
\newcommand{\given}{\,\,\big\vert\,\,}

\newcommand{\ntr}{n_{ \rm{tr} }}
\newcommand{\ncal}{n_{ \rm{cal} }}

\newcommand{\Dtr}{\mathcal{D}_{ \rm tr }}
\newcommand{\Dcal}{\mathcal{D}_{ \rm cal }}

\newcommand{\Lmax}{\bar{L}} 
\newcommand{\Rmax}{\bar{R}}
\newcommand{\Wmax}{\bar{W}}
\newtheorem{proposition}{Proposition} 
\newtheorem{assumption}{Assumption}  
  
\newtheorem{definition}{Definition} 
\newtheorem{lemma}{Lemma}
\newtheorem{theorem}{Theorem}
\begin{document}
\title{Generalization and Informativeness of Weighted Conformal Risk Control Under Covariate Shift}

	\author{%
  \IEEEauthorblockN{Matteo Zecchin$^\dag$, Fredrik Hellström$^{*}$, Sangwoo Park$^\dag$,  Shlomo Shamai (Shitz)$^\mathsection$, and Osvaldo Simeone$^\dag$}
  \IEEEauthorblockA{$^\dag$KCLIP lab, Centre for Intelligent Information Processing Systems (CIIPS), King’s College London, London, UK}
  \IEEEauthorblockA{$^{*}$Centre for Artificial Intelligence, University College London, London, UK}
  \IEEEauthorblockA{$^\mathsection$Faculty of Electrical and Computing Engineering, Technion—Israel Institute of Technology, Israel}}

\maketitle

\begin{abstract}
    Predictive models  are often required to produce reliable predictions under statistical conditions that  are not matched to the training data. A common type of training-testing mismatch is covariate shift, where the conditional distribution of the target variable given the input features remains fixed, while the marginal distribution of the inputs changes. Weighted conformal risk control (W-CRC) uses data collected during the training phase to convert point predictions into prediction sets with valid risk guarantees at test time despite the presence of a covariate shift. However, while W-CRC provides statistical reliability, its efficiency -- measured by the size of the prediction sets -- can only be assessed at test  time.
    In this work, we relate the generalization properties of the base predictor to the efficiency of W-CRC under covariate shifts. 
    Specifically, we derive a bound on the inefficiency of the W-CRC predictor that depends on algorithmic hyperparameters and task-specific quantities available at training time. This bound offers insights on relationships between the informativeness of the prediction sets, the extent of the covariate shift, and the size of the calibration and training sets. Experiments on fingerprinting-based localization validate the theoretical results.
\end{abstract}
\section{Introduction}

\label{sec:intro}

In many applications, the dependability of prediction models relies on their ability to quantify the uncertainty in their outputs \cite{hewing2020learning,lu2022fair,angelopoulos2024theoretical,zecchin2024forking}. A significant challenge in producing reliable uncertainty quantification is the presence of \emph{distribution shifts} between training and testing conditions, which can lead to mismatched and unreliable estimates \cite{kuleshov2017estimating,ovadia2019can}. A particular class of distribution shifts is  \emph{covariate shift}, where the distribution of the prediction target given the input covariates remains fixed between training and testing,  while the marginal distribution of the covariates is allowed to change \cite{shimodaira2000improving}.

\begin{figure}[t]
    \centering
    \includegraphics[width=\linewidth]{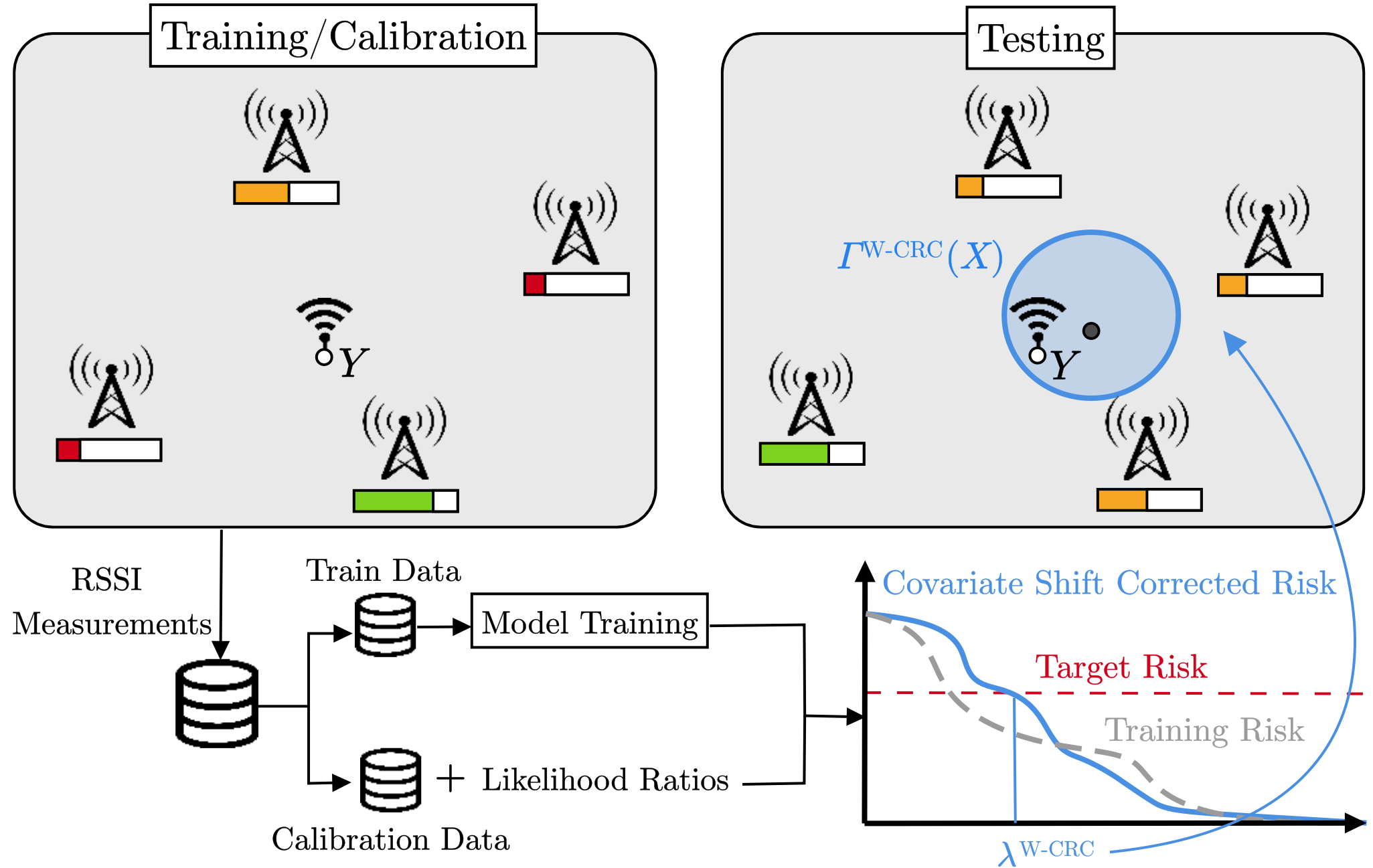}\vspace{-0.25em} 
    \caption{In fingerprinting-based localization, received signal strength indicators (RSSI)  measured by base stations are used to localize mobile devices. The base stations can be dynamically switched off, and their activation rates are  depicted by the colored bars. A portion of the collected RSSI measurements is used to train a prediction model, while the rest are used to convert the model's outputs into prediction regions that meet a specified target risk. Calibration is achieved by W-CRC by accounting for covariate shift due to differing activation rates during training and testing through likelihood ratios between the training and testing marginals \cite{tibshirani2019conformal,angelopoulos2022conformal}.}
    \label{fig:motivating_example}
    \vspace{-1.5em}
\end{figure}

To illustrate the problem of interest, consider the task of fingerprinting-based localization in cellular systems \cite{yiu2017wireless,pecoraro2018csi}, depicted in Figure \ref{fig:motivating_example}. In this scenario, measurements and transmitter locations are collected for training under specific base station  activation patterns, whereby base stations are dynamically switched off to save energy. However, cellular  operators may need reliability estimates under different base station activation conditions depending on the conditions encountered at test time \cite{feng2017base}.

\emph{Weighted  conformal risk control} (W-CRC) transforms a potentially unreliable predictor into a calibrated set predictor  \cite{tibshirani2019conformal,angelopoulos2022conformal,angelopoulos2023conformal}. For example, in fingerprinting-based localization, W-CRC converts point estimates of a mobile's location into geographical uncertainty regions. The prediction sets come  with assumption-free risk guarantees -- e.g., on the coverage probability for the localization task -- that hold irrespective of the covariate shift. 

As illustrated in Figure \ref{fig:motivating_example},  W-CRC  first splits the available data into training and calibration sets. The training set is used to train a base prediction model, while the calibration set is used to construct error bars -- more generally, a prediction set -- around the model's outputs. To account for a covariate shift, W-CRC leverages knowledge of the likelihood ratio of the training and test marginal distributions.

While the prediction sets produced by W-CRC guarantee a user-defined risk level, the \emph{informativeness} of the prediction sets can only be evaluated at test time. In this context, the goal of this work is to relate the generalization properties of the base predictor to the average size of the set predictor produced by W-CRC, where larger prediction sets are less informative.

In the absence of distribution shift between training and testing conditions, the expected size of conformal prediction sets \cite{angelopoulos2024theoretical} has been analyzed \cite{dhillon2024expected} and connected to the generalization performance of the base predictor in the finite-sample regime \cite{matteo2024informativeness}. Prior works studied optimal prediction sets with minimal size in the asymptotic regime \cite{lei2018distribution}. Also related are works with an algorithmic focus that aim to increase prediction efficiency by optimizing the scoring function \cite{yan2024provably, zargarbashi2024robust}.  

In this paper, after defining the problem (Section II), we provide a bound on the average prediction set size that depends  on algorithmic hyperparameters and task-dependent quantities available at training time (Section III). This result offers insights into the relationship between informativeness, the extent of the covariate shift, reliability requirements, and the amount of calibration and training data. Experimental evidence based on a real-world fingerprinting localization task validates the conclusions derived from the analysis (Section IV).

\section{Problem Definition}

\label{sec:problem_def}

\subsection{Setting}

We consider a supervised learning setting, where a data point $z = (x, y) \in \mathcal{X} \times \mathcal{Y} = \mathcal Z$ consists of an input feature $x \in \mathcal{X}$ and a label $y \in \mathcal{Y}$.  The learner has access to a data set $\mathcal{D} = \{Z_i\}_{i=1}^{n}\in \mathcal Z^n$  consisting of independent and identically distributed (i.i.d.) pairs $Z_i=(X_i,Y_i)$ following  an \emph{unknown training distribution} $P_{XY}$. The test data $Z=(X, Y)$ is sampled from an \emph{unknown test distribution}  $P'_{XY}$. 

Specifically, reflecting a  \emph{covariate shift},  the test distribution  $P'_{XY}=P'_X P'_{Y|X}$ has the same conditional label distribution as the training distribution $P_{XY}=P_X P_{Y|X}$, but differs in the input distribution.
That is, we have the conditions  $P'_{Y|X} = P_{Y|X}$ and $P'_X\neq P_X$ \cite{shimodaira2000improving}.

For a test data point $Z=(X, Y)$, a set predictor $\Gamma$ outputs a subset of the label space $\Gamma(X) \subseteq \mathcal{Y}$. The quality of the set predictor is evaluated through a bounded loss function $\mathcal{L}(\Gamma(X), Y)\in[0,\Lmax]$. One popular loss function is the miscoverage loss
   \begin{equation}
   \label{eq:mis_loss}
   \mathcal{L}(\Gamma(X), Y) = \mathbbm{1}\{Y \notin \Gamma(X)\},
   \end{equation}
which equals $1$ if the true label $Y$ is not included in the set  $\Gamma(X)$, and equals  $0$ otherwise, yielding $\bar{L}=1$. Another common choice is the point-to-set distance
\begin{equation}\label{eq:l_2_set_loss}
    	\mathcal{L}(\Gamma(X),Y) = \min_{Y' \in \Gamma(X)} \lVert Y - Y' \rVert_p,  
\end{equation}
which corresponds to the Minkowski distance of order $p$ between the predicted set $\Gamma(X)$ and the label $Y$. If the target domain is bounded, i.e. there exists $B$ such that $\lVert y \rVert_p\leq B$ for all $y\in\mathcal{Y}$, the point-to-set distance yields $\Lmax=2B$ \cite{scholkopf1998shrinking}.
When the label $Y$ itself is a set, one can also consider loss functions such as the false negative rate or the  F1-score \cite{angelopoulos2022conformal}.

For a given loss function, the set predictor $\Gamma$ is said to be $\alpha$\emph{-reliable} if the average risk under the test distribution does not exceed a user-defined threshold $\alpha\in [0,\Lmax]$, i.e., \begin{align}  
	\label{eq:crc_guarantee}
	\mathbb{E}
    [\mathcal{L}(\Gamma(X),Y)] \leq \alpha 
\end{align}
where the expectation is over the test point $(X, Y)\sim P'_{XY}$.
 
Most loss functions used to evaluate set predictors, including  the miscoverage loss (\ref{eq:mis_loss}) and the point-to-set loss (\ref{eq:l_2_set_loss}), decrease with the size of the prediction set $\Gamma(X)$ and are zero when $\Gamma(X)=\mathcal{Y}$. This condition is formalized by the following assumption \cite{tibshirani2019conformal,angelopoulos2022conformal,angelopoulos2023conformal}.

\begin{assumption}[Bounded, non-increasing loss]
\label{ass:loss_bound_noincr}
There exists a constant $\Lmax<\infty$ such that $\mathcal{L}(\Gamma(x), y)\in[0,\Lmax]$ for any pair $(x, y)\in\mathcal Z$. Moreover, if $\Gamma'(x)\subseteq \Gamma(x)$, we have the inequality \begin{equation}\mathcal{L}(\Gamma(x), y)\leq \mathcal{L}(\Gamma'(x), y)\end{equation} for any $y\in \mathcal{Y}$. 
 In particular, with $\Gamma(x)=\mathcal{Y}$, we have \begin{equation}\label{eq:equalzero}\mathcal{L}(\mathcal{Y}, y)=0.\end{equation}
\end{assumption}

The condition (\ref{eq:equalzero})  implies  that the reliability guarantee \eqref{eq:crc_guarantee} can be ensured by a set predictor that always returns the entire label space, i.e., $\Gamma(x) = \mathcal{Y}$ for all $x\in\mathcal X$. Therefore, to quantify the quality of a set predictor, it  is critical to evaluate its \emph{inefficiency}, which is defined as the expected size of the predicted set with respect to the test data $X\sim P'_{X}$:
\begin{align}
    \label{eq:inefficiency}
    \Lambda(\Gamma) = \mathbb{E}[|\Gamma(X)|].
\end{align}

\subsection{Training}

We partition the data $\mathcal{D}$ into a training set $\Dtr$, used to train a base predictor, and a calibration set $\Dcal=\mathcal{D}\setminus \Dtr$ that is used to calibrate the predictor's outputs to ensure the $\alpha$-reliability condition (\ref{eq:crc_guarantee}).  
We denote the size of $\Dtr$ as $\ntr$ and the size of $\Dcal$ as $\ncal$.
Specifically, we set $(X_i, Y_i)\in \Dcal$ for $i\in \{1,\dots,\ncal\}$, while $(X_i, Y_i)\in \Dtr$ for $i\in\{\ncal+1,\dots,n\}$.

Given a model class $\mathcal{F}=\{f_\theta:\theta\in\Theta\}$, where $f_\theta:\mathcal{X}\to\predspace$ is a point predictor parameterized by a vector $\theta$, the training data $\Dtr$ is used to run a learning algorithm that produces a probability distribution $Q(\theta|\Dtr) $ over the space of model parameters $\Theta$. Examples of learning algorithms include stochastic gradient descent for conventional frequentist learning \cite{robbins1951stochastic,welling2011bayesian}, as well as  variational inference algorithms for Bayesian learning \cite{bishop2006pattern,simeone2022machine}.

For an input $x$, predictions are obtained using Thompson sampling: A parameter $ \theta\sim Q(\theta|\Dtr)$ is first sampled, and the resulting model is then used to obtain a prediction $\hat{y}=f_\theta(x)$.

The \emph{generalization error} of a learning algorithm $Q(\theta|\Dtr) $ is the difference between the performance of the predictor during training and testing. In the case of  a covariate shift, at training time, the performance of the predictor is evaluated using samples from the training distribution $P_{XY}$, while at test time it is assessed using samples from the test distribution $P'_{XY}$. Thus, the generalization error of a learning algorithm depends on both the training set size $\ntr$ and the extent of the covariate shift \cite{blitzer2007learning,mansour2009domain}.

\subsection{Weighted Conformal Risk Control}

Given a test input $x$, W-CRC augments the point prediction $f_\theta(x)$ with a set predictor $\Gamma_\theta(x, \lambda)$. To this end, W-CRC relies on the evaluation of a  \emph{non-conformity (NC) scoring function} $R\!:\predspace\times\mathcal{Y}\to [0, \Rmax]$ that evaluates the mismatch of the predictor $f_\theta(x)$ with respect to label $y$ as $R(f_\theta(x), y)$. For example, a typical scoring function for regression is the squared loss $R(f_\theta(x), y)=(y-f_\theta(x))^2$. The set predictor is parameterized by a threshold  $\lambda\in[0, \Rmax]$, and  it includes all the labels $y\in\mathcal{Y}$ whose NC score does not exceed the threshold $\lambda$:
\begin{align}
    \Gamma_\theta(x, \lambda) = \{y \in \mathcal{Y} : R(f_\theta(x), y) \leq \lambda\}.
    \label{eq:set_predictor}
\end{align}
The threshold $\lambda$ controls the size of the set predictor \eqref{eq:set_predictor}. In W-CRC, the value of $\lambda$ is determined based on the calibration set $\Dcal$ and on information about the covariate shift.

The covariate shift determines  the \emph{likelihood ratios} 
\begin{align}\label{eq:likratio}
    W_i = w(X_i) = \frac{P_X'(X_i)}{P_X(X_i)} 
\end{align} for all the calibration data points  $i\in \{1,\dots,\ncal\}$. Denote as \begin{align}\label{eq:calloss}
    L_i(\lambda) = \mathcal{L}(\Gamma_{\theta_i}(X_i, \lambda), Y_i) 
\end{align} the loss of the set predictor for each calibration point $(X_i, Y_i)$ with $i\in \{1,\dots,\ncal\}$, where  $\{\theta_i\}_{i=1}^n\cup \theta$ are drawn i.i.d. from $Q(\theta|\Dtr)$ following Thompson sampling. 
Note that this implementation  deviates slightly from previous versions of W-CRC, which use a fixed (deterministic) point predictor for all samples \cite{angelopoulos2022conformal}. 
Based on the calibration loss values (\ref{eq:calloss}) and likelihood ratios (\ref{eq:likratio}), given the test input $X$, W-CRC evaluates the following empirical estimate of the calibration loss 
\begin{equation}\label{eq:weighted_aug_risk}
\!\!\hat{\mathcal{L}}_Q(\lambda, X |\Dfull) = \frac{1}{\sum_{i=1}^{\ncal} W_i \!+\! W} \left( \sum_{i=1}^{\ncal} W_i L_i(\lambda) \!+\! W \Lmax \!\right),
\end{equation}
where $W = w(X)$ denotes the likelihood ratio evaluated at the test data point $X$. The loss \eqref{eq:weighted_aug_risk} can be interpreted as the covariate-shift corrected empirical loss of the set predictor  when evaluated on the calibration set augmented with a test point $X$ with maximal loss value $\Lmax$.

For a user-defined reliability level $\alpha\in[0,\Lmax]$ in the constraint (\ref{eq:crc_guarantee}), W-CRC evaluates the threshold $\lambda$ in \eqref{eq:set_predictor} as
\begin{align}
    \label{eq:crc_th}
    \lambdawcrc(X|\Dfull) = \inf\{\lambda : \hat{\mathcal{L}}_Q(\lambda, X | \Dfull) \leq \alpha\}.
\end{align} Accordingly, the selected threshold is the smallest value of $\lambda$ that ensures an empirical estimate (\ref{eq:weighted_aug_risk}) not exceeding the target $\alpha$. This 
yields the set predictor
\begin{equation}
    \!\!\!\Gammawcrc(X\vert \Dfull) \!=\! \{y \!\in\! \mathcal{Y} \!: R(f_\theta(X), y) \!\leq\! \lambdawcrc(X|\Dfull)\} \!\label{eq:crc_set_predictor}
\end{equation}
with the sampled model parameter $\theta \sim Q(\theta|\Dtr)$.

For loss functions satisfying Assumption \ref{ass:loss_bound_noincr}, W-CRC produces $\alpha$-reliable prediction sets.
\begin{proposition}[Proposition 4\cite{angelopoulos2022conformal}]
    \label{prop:crc_weighted}
    Under Assumption \ref{ass:loss_bound_noincr}, the W-CRC predictor \eqref{eq:crc_set_predictor} is $\alpha$-reliable, i.e.
    \begin{align}
        \label{eq:crc_guarantee_prop}
        \mathbb{E}[\mathcal{L}(\Gamma^{\rm W\text{-}CRC}_{\theta}(X \vert \Dfull), Y) \given \Dtr]\leq \alpha,
    \end{align}
    where the expectation is over both the test and calibration data as well as over the model parameters, with fixed training set $\Dtr$.  This result follows in a manner similar to \cite{tibshirani2019conformal,angelopoulos2022conformal}, with the only caveat that one needs to account for the random choice of model parameters via Thompson sampling. For completeness, a proof is provided in Appendix A.

\end{proposition}

\section{Information-Theoretic Inefficiency Bound for W-CRC}

As seen in the previous section, W-CRC ensures the reliability guarantee (\ref{eq:crc_guarantee}) for the set predictor  $\Gammawcrc(X \vert \Dfull)$. However, the \emph{inefficiency} 
\begin{align}
 	\Lambdawcrc=\mathbb{E}\big[|\Gammawcrc(X \vert \Dfull)| \given \Dtr\big]
    \label{eq:ineff_w_crc}
\end{align} can generally only be evaluated at test time. In this section, we derive theoretical bounds on the inefficiency (\ref{eq:ineff_w_crc}).

\begin{figure}[t]
    \centering
    \includegraphics[width=0.75\linewidth]{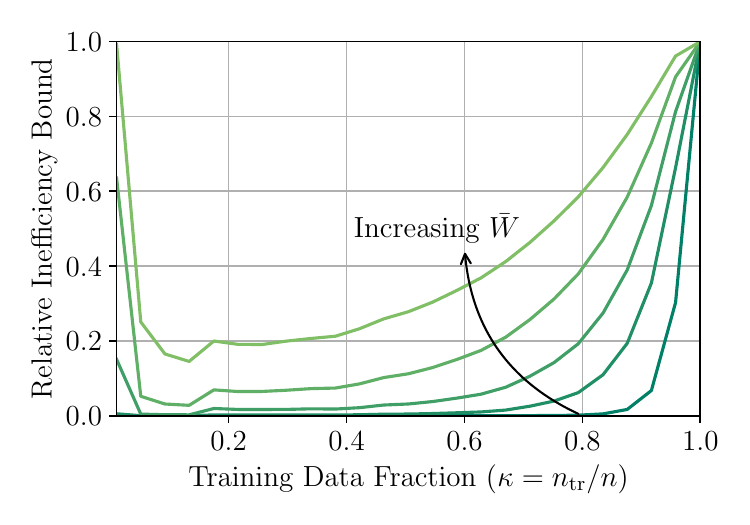}
    \vspace{-1.2em}
    \caption{Inefficiency bound as a function of the fraction  $\kappa$ of data allocated for training under a fixed data budget $n$ while varying the covariate shift parameter $\bar{W}$. }
    \label{fig:bound_eval}
    \vspace{-1.5em}
\end{figure}

\subsection{Generalization Properties of the Base Predictor}
The bound provided in this work relates the generalization properties of the base predictor to the informativeness of the set predictor produced by W-CRC. Accordingly, to start, we make a standard assumption on the behavior of the generalization gap of the training algorithm. 

Specifically, we consider  the generalization properties of the base predictor $Q(\theta|\Dtr)$ in terms of the risk $\mathcal{L}(\cdot,\cdot)$ of the set predictor \eqref{eq:set_predictor}. The generalization gap accounts for the difference between training and test risk measures. 
 The training risk of the set predictor with hyperparameter $\lambda$ is defined as the following sum over the training data points
\begin{align}
    \hat{\mathcal{L}}_Q(\lambda|\Dtr)=\frac{1}{\ntr}\sum^{n}_{i=\ncal+1}\!\!\!\mathbb{E}\left[\mathcal{L}(\Gamma_{\theta_i}(X_i,\lambda),Y_i) \given \Dtr \right], 
    \label{eq:train_risk_estimate}
\end{align}
while the corresponding population risk under the test distribution is given by
\begin{align}
    \mathcal{L}_Q(\lambda)=\mathbb{E}\left[\mathcal{L}(\Gamma_{\theta}(X,\lambda),Y)\right] .
     \label{eq:test_risk_estimate}
\end{align} Note that the average in (\ref{eq:train_risk_estimate}) is with respect to the model parameters, while in (\ref{eq:test_risk_estimate}) the average is also over the test distribution $P'_{XY}$.
As generalization metric, we adopt the maximum absolute difference between the training risk estimate \eqref{eq:train_risk_estimate} and the population test risk \eqref{eq:test_risk_estimate} over the hyperparameter $\lambda$:
\begin{align}
    \Delta(Q|\Dtr)=\sup_{\lambda\in[0,\Rmax]}|\hat{\mathcal{L}}_Q(\lambda|\Dtr)-\mathcal{L}_Q(\lambda)|.
    \label{eq:gen_metric}
\end{align}

With this definition, we make the following assumptions.

\begin{assumption}[Bounded likelihood ratio]
    \label{ass:bound}
    There exists a constant $\Wmax<\infty$ such that the likelihood ratio is upper bounded as $w(x)\leq \Wmax$ for all $x\in\mathcal{X}$.
\end{assumption} 

This assumption implies that the parameter $\Wmax$ provides a measure of the extent of the covariate shift.

\begin{assumption}
    \label{ass:gen_base}
    For any $\delta\in(0,1)$, there exists a function $\beta(\delta,\ntr)=O(\log(\ntr))$ such that, with probability at least $1-\delta$ with respect to the training set $\Dtr\sim P_{XY}^{\otimes \ntr}$, we have
    \begin{align}
        \label{eq:gen_bound}
        \Delta(Q|\Dtr)\leq \frac{\beta(\delta,\ntr)}{\sqrt{\ntr}}+\Lmax\sqrt{\log(\Wmax)}.
    \end{align}
\end{assumption}
By Assumption \ref{ass:gen_base}, the difference between the empirical risk under $P_{XY}$ and the population risk under the test distribution $P'_{XY}$ can be bounded by a term that vanishes   in the size of the training set,  $\ntr$,  with the addition of  an irreducible bias term caused  by the presence of a  covariate shift. It is worth noticing that the generalization metric \eqref{eq:gen_bound}  bounds the risk difference uniformly over the hyperparameter $\lambda$. In Appendix \ref{app:gen_base_pred}, we prove that Assumption \ref{ass:gen_base} is satisfied for many popular learning algorithms, such as Gibbs posteriors \cite{bishop2006pattern}, $\epsilon$-differentially private algorithms with $\epsilon=O(\log(\ntr)/\ntr)$ \cite{chaudhuri2011differentially}, and stochastic gradient Langevin dynamics (SGLD) \cite{welling2011bayesian}.

\subsection{Analysis of the NC Score}

In order to derive bounds on the inefficiency of the W-CRC set predictor, an important quantity to study is the size of the NC score $R(\cdot,\cdot)$ \cite{dhillon2023expected}. In fact, the prediction set (\ref{eq:set_predictor}) includes all labels with an NC score smaller than threshold $\lambda$.

To this end, we let $P_{R|Y=y,\theta}$ denote the probability density function of the NC score $R(f_\theta(X),y)$ for a given model $\theta$ under the \emph{test} distribution $P'_{X|Y=y}=P'_{X,Y=y}/P'_{Y=y}$. Hence, $P_{R|Y=y,\theta}$ 
 indicates how likely an NC score value is given the model $\theta$ and the label $y$. Averaging over the sampled model $\theta\sim Q(\theta|\Dtr)$ and a uniformly chosen label $Y\sim\mathcal{U}(\mathcal{Y})$, the \emph{size of an NC score value $r$} \cite{dhillon2023expected} is defined as
\begin{align}
    \gamma(r|Q,\Dtr)=\frac{1}{|\mathcal{Y}|}\int_{\mathcal{Y}}  \mathbb{E}
    \left[P_{R|Y=y,\theta}(r) \given \Dtr \right]\mathrm{d}y.
    \label{eq:exp_mf}
\end{align}
To streamline notation, we denote $\gamma(r|Q,\Dtr)$ as $\gamma(r)$. A larger $\gamma(r)$ indicates that the NC score $r$ is more likely.

\subsection{Main Result}

With the assumptions listed above, we obtain the following guarantee on the informativeness of the W-CRC set predictor, the proof of which is provided in Appendix \ref{app:main_thm_pf}.
\begin{theorem}
    \label{th:gen_bound}
    Suppose that Assumptions \ref{ass:loss_bound_noincr}-\ref{ass:gen_base} hold.
    Then, with probability at least $1-\delta$ with respect to the training set $\Dtr\sim P_{XY}^{\otimes \ntr}$, the expected set size of the W-CRC predictor satisfies the inequality 
    \begin{align}\label{eq:gen_bound_W}
       &\frac{\Lambdawcrc}{|\mathcal{Y}|}\leq\int^{\hat{\lambda}}_{0} \gamma(r)\mathrm{d}r+ \\&\hspace{-0.5em}+\hspace{-0.25em}\int^{\Rmax}_{\hat{\lambda}}	\hspace{-0.75em}e^{-\frac{2\ncal}{\Lmax^2\Wmax^2}\left(\!\frac{\Wmax(\alpha-\Lmax)}{\ncal}-
		\hat{\mathcal{L}}_Q(r|\Dtr)+\alpha-\frac{\beta(\delta,\ntr)}{\sqrt{\ntr}}-\Lmax\sqrt{\log(\Wmax)}\right)^2} \!\!\gamma(r)\mathrm{d}r,
        \nonumber
    \end{align}
    where the value $\hat{\lambda}$ is defined as
    \begin{multline}
	\hat{\lambda}=\inf \Bigg\{r:	\hat{\mathcal{L}}_Q(r|\Dtr)\leq\\
    \alpha+\frac{\Wmax(\alpha-\Lmax)}{\ncal}-\frac{\beta(\delta,\ntr)}{\sqrt{\ntr}}-\Lmax\sqrt{\log(\Wmax)}\Bigg\}.
    \end{multline}
\end{theorem}

\subsection{Discussion}
Analyzing the bound in Theorem \ref{th:gen_bound}, it is possible to study the interplay between the efficiency of W-CRC, the extent of the covariate shift, and the sizes of the calibration and training sets. In particular, we make the following observations:

\noindent $\bullet$ A larger covariate shift, reflected in a larger  maximum likelihood ratio $\Wmax$,  causes  the inefficiency of W-CRC to increase. This is a consequence of the negative impact of a covariate shift on the generalization properties \eqref{eq:gen_bound} of the base predictor.  

\noindent $\bullet$ A larger size of the training set, $\ntr$,  improves the performance of the base predictor, as reflected in a smaller value of the generalization error bound (\ref{eq:gen_bound}). However,  for large values of the covariate shift measure $\Wmax$, increasing the value of $\ntr$ has a limited effect on reducing the inefficiency, and the effect of covariate shift dominates.

 \noindent $\bullet$  A larger covariate shift  reduces the decay rate of the exponential function  in the second term of the bound \eqref{eq:gen_bound_W}. Given the dependence of this term on the calibration set size $\ncal$, this result  suggests that a larger calibration set is required in order to counteract the effect of a covariate shift.

\noindent $\bullet$ More broadly, a larger  calibration set size $\ncal$ reduces the inefficiency bound for a fixed training set size and covariate shift. A larger calibration set enables a sharper estimation of the CRC threshold, reducing the impact of covariate shift. This improvement is reflected both in the exponential term of \eqref{eq:gen_bound_W}, which decays faster as $\ncal$ grows, and in the quantity  $\hat{\lambda}$ which decreases as $\ncal$ grows.

Overall, the bound suggests that the performance of W-CRC is negatively affected by covariate shift, especially when limited calibration data is available. Accordingly, in settings with severe covariate shift, it may be beneficial to allocate a larger portion of the data for calibration. This is illustrated in Figure \ref{fig:bound_eval}, where the bound is evaluated for a fixed data budget $n$ with varying training data fraction $\kappa=\ntr/n$. It is observed that an optimized data split that minimizes the inefficiency of W-CRC shifts in favor of larger calibration sets as the covariate shift becomes more significant, and thus  $\Wmax$ increases.
\begin{figure*}
\centering
\begin{minipage}{.55\textwidth}
  \centering
  \includegraphics[width=\linewidth]{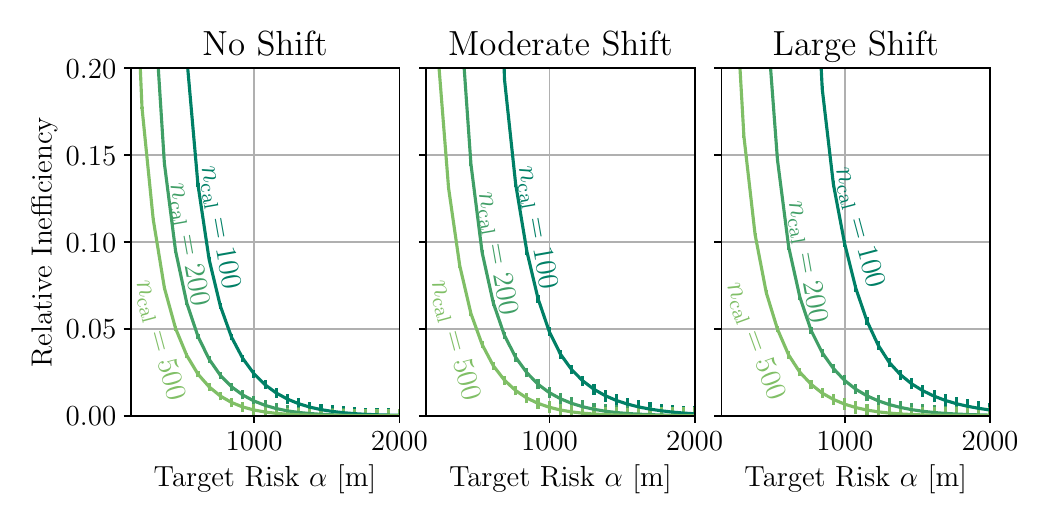}
    \vspace{-2em}
   \caption{Relative inefficiency of W-CRC as a function of the target reliability level $\alpha$ for different calibration set sizes $\ncal$ and various covariate shift settings.}
    \label{fig:rssi_loc_1}
     \vspace{-1.5em}
\end{minipage}%
\hfill
\begin{minipage}{.43\textwidth}
  \centering
  \includegraphics[width=.91\linewidth]{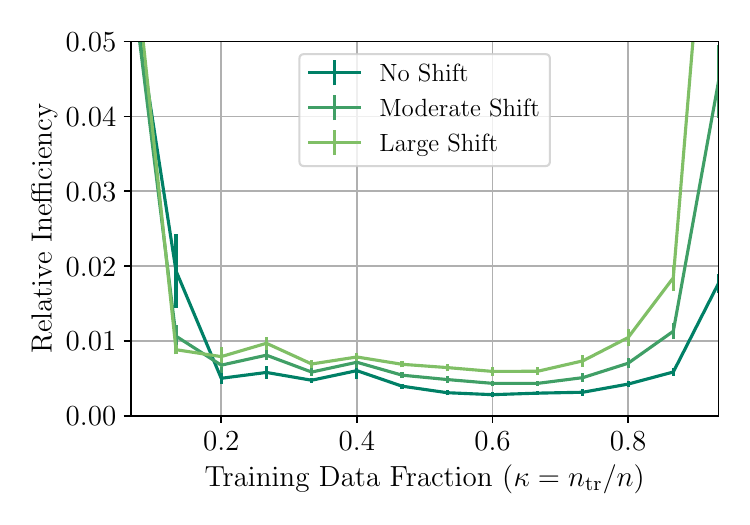}
  \vspace{-1em}
  \caption{Relative inefficiency of W-CRC as a function of the fraction of training data $\kappa$ under a fixed data set of size $n$ and various covariate shift settings.  }
    \label{fig:rssi_loc_2}
     \vspace{-1.5em}
\end{minipage}
\end{figure*}

\section{Experiments}

In this section, we consider the RSSI-based localization problem introduced in Section \ref{sec:intro}. The experiments use data from the rural Sigfox data set \cite{data3020013}, which contains real-world RSSI measurement vectors $X_{\rm RSSI} \in \mathbb{R}^d$ collected from $d = 137$ base stations (BSs) located between the cities of Antwerp and Ghent.

We assume that, during the operation of the system, a fixed subset $\mathcal{B}$ of the BSs can be switched off \cite{feng2017base}, and this event is denoted by the random variable $X_{\rm act} = \mathbbm{1}\{\mathcal{B} \text{ is inactive}\}$. When a BS in $\mathcal{B}$ is deactivated, the corresponding RSSI measurement is set to zero. In this setup, we introduce a probability $p_{\rm off}$ of deactivating the BSs in $\mathcal{B}$ during training, and a possibly different probability $p'_{\rm off}$ during testing. Specifically, we assume $p_{\rm off} = 0.5$ during training and consider three possible values for $p'_{\rm off}\in \{0.5, 0.75, 0.9\}$ during testing. Accordingly, this setup represents three scenarios: no covariate shift ($p_{\rm off} = p'_{\rm off} = 0.5$), moderate covariate shift ($p'_{\rm off} = 0.75$), and large covariate shift ($p'_{\rm off} = 0.9$).

Based on the RSSI measurements $X_{\rm RSSI}$ and the BS activation pattern $X_{\rm act}$, the goal is to reliably estimate the transmitter’s location $Y \in \mathbb{R}^2$. To achieve this, we apply W-CRC to calibrate a localization model $f_\theta(X)$, realized via a 3-layer Bayesian neural network with parameter vector $\theta \sim Q(\theta|\Dtr)$. The posterior $Q(\theta|\Dtr)$ is given by a parameterized variational distribution optimized using stochastic gradient descent.

Based on the predicted location $\hat{Y} = f_\theta(X)$, we consider a W-CRC set predictor in the form \eqref{eq:set_predictor} based on the NC score
\begin{align}
    R(f_\theta(X), Y) = \lVert f_\theta(X) - Y \rVert_2.
\end{align}
With this NC score, the predicted sets are circular geographical regions that are centered at the prediction $f_\theta(X)$ and have a radius determined by the W-CRC hyperparameter \eqref{eq:crc_th}.

The set predictor radius is calibrated to ensure $\alpha$-reliability with respect to the loss \eqref{eq:l_2_set_loss} with $p = 2$. This way, the loss is proportional to the Euclidean distance between the predicted region and the ground truth location.

In Figure \ref{fig:rssi_loc_1}, we analyze the W-CRC performance under the three covariate shift settings by plotting the set predictor’s inefficiency, normalized by the deployment area, against the reliability level, $\alpha$. The base model is trained using $\ntr = 1000$ data points and we consider three calibration set size values $\ncal\in\{100,200,500\}$. For all covariate shift settings, the relative inefficiency is seen to  increase as the target risk $\alpha$ is reduced. Moreover, as predicted by the inefficiency bound in Theorem \ref{th:gen_bound}, the impact of covariate shift is more significant with smaller calibration sets. For example, for $\ncal = 500$, the inefficiency curve remains similar across the covariate shift settings, while it  degrades significantly for  $\ncal = 100$.

In the previous experiment, we varied the calibration set size $\ncal$ under a fixed training data budget, $\ntr$. Here, we consider a fixed total data set of size $n = 1500$ and vary the fraction $\kappa = \ntr / n$ of data allocated for training as in Figure 2. This approach allows us to investigate the performance of W-CRC under different training-calibration splits. In Figure \ref{fig:rssi_loc_2}, we report the relative inefficiency of W-CRC as a function of the training data fraction $\kappa$ under the three covariate shift scenarios and a fixed reliability level $\alpha=1$ km. We vary $\kappa$ from $1/15$, in which case $\ntr = 100$ and $\ncal = 1400$, to $14/15$, in which case $\ntr = 1400$ and $\ncal = 100$. The performance of W-CRC deteriorates for extreme values of $\kappa$. At low $\kappa$, this is due to the poor generalization of the base predictor, while for large $\kappa$ it is due to an insufficient amount of calibration data to estimate the W-CRC threshold \eqref{eq:crc_th}.

The results confirm the intuition provided by the inefficiency bound and highlight the increased importance of calibration data under severe covariate shifts. In fact, as the extent of covariate shift increases, data splits for which W-CRC performance remains satisfactory shift toward values that favor using more calibration data.

\section{Conclusion}
In this work, we have considered the problem of calibrating a predictive model under covariate shift by studying the inefficiency of the state-of-the-art calibration scheme W-CRC. Specifically, we have derived a bound on the expected size of the prediction sets that are linked to the generalization properties of the base predictor, the magnitude of covariate shift, and algorithmic parameters such as the size of the training and calibration sets. 
The derived bound provides insight into sensible training-calibration splits, and on how the optimal split varies depending on the severity of covariate shift.
Extending these results to general distribution shifts could be a promising direction for future research.

\section*{Acknowledgments}
This work was partially supported by the European Union’s Horizon Europe project CENTRIC (101096379), by the Open Fellowships of the EPSRC (EP/W024101/1), by the EPSRC project (EP/X011852/1), and by the Wallenberg AI, Autonomous Systems and Software Program (WASP) funded by the Knut and Alice Wallenberg Foundation. The work of S. Shamai was supported by the German Research Foundation (DFG) via the German-Israeli Project Cooperation (DIP), under Project SH 1937/1-1, and by the Ollendorff Minerva Center of the Technion.

\bibliographystyle{IEEEtran}
\bibliography{ref}

\newpage
\begin{appendix}
    \subsection{Proof of Proposition \ref{prop:crc_weighted}}
    \label{app:crc_risk_guar}
    In the following, we prove that the W-CRC set predictor satisfies the $\alpha$-reliability requirement, as stated in Proposition \ref{prop:crc_weighted}, under the assumption that the calibration data $\Dcal$ and the test data point $(X,Y)$ are weighted exchangeable \cite{tibshirani2019conformal}.
    \begin{definition}[Weighted Exchangeability \cite{tibshirani2019conformal}]
        The random variables $X_1,\dots,X_{m}$ are weighted exchangeable, with weight functions $w_1(\cdot),\dots,w_m(\cdot)$, if the joint distribution $P(X_1,\dots,X_m)$ can be factorized as 
        \begin{align}P(X_1,\dots,X_m)=\prod^m_{i=1}w_i(X_i)g(X_1,\dots,X_m)
        \end{align}
        where $g(\cdot,\dots,\cdot)$ does not depend on the ordering of its inputs.
    \end{definition}
    
    Note that independent random variables that have a density with respect to a common measure are weighted exchangeable.
    Hence, the proof applies directly to the data model considered in this work.
    
    Denote the risk of the set predictor $\Gamma_\theta(\cdot,\lambda)$ with model $\theta\sim Q(\theta|\Dtr)$, evaluated at an unknown test data point $(X, Y) \sim P'_{XY}$, as $L_{\ncal+1}(\lambda)$ and the associated likelihood ratio $W(X)$ as $W_{\ncal+1}$. 
    
    The weighted empirical risk, evaluated on the augmented data set $\Dcal’ = \Dcal \cup (X, Y)$, is given by
   \begin{align}
        \label{eq:cal_test_risk}
        \hat{\mathcal{L}}_Q(\lambda |\Dcal') = \frac{1}{\sum_{i=1}^{\ncal+1} W_i}\hspace{-0.5em}\sum_{i=1}^{\ncal+1} \hspace{-0.5em} W_i L_i(\lambda).
    \end{align}
    Note that the risk is a function of the data $\Dcal'$ as well as the sampled models $\{\theta_i\}^{\ncal}_{i=1}\cup\{\theta\}$. 
    The augmented weighted empirical calibration risk \eqref{eq:weighted_aug_risk} upper bounds the risk \eqref{eq:cal_test_risk} since, from Assumption \ref{ass:loss_bound_noincr}, $L_{\ncal+1}(\lambda)\leq \Lmax$ for all values of $\lambda$. 

    Based on the empirical risk \eqref{eq:cal_test_risk}, we define the threshold
    \begin{align}
        \label{eq:fic_lambda}
        \lambda'=\inf\{\lambda: \hat{\mathcal{L}}_Q(\lambda |\Dcal')\leq \alpha\}.
    \end{align}
    Given that the loss is non-increasing in $\lambda$, we have $\lambda'\leq \lambdawcrc$ in \eqref{eq:crc_th} and
    \begin{align}
        \label{eq:risk_ineq_aug}
         \mathbb{E}[L_{\ncal+1}(\lambdawcrc)]\leq  \mathbb{E}[L_{\ncal+1}( \lambda')].
    \end{align}
    Let $E$ be the multiset of data and model pair realizations $\{(Z_1,\theta_1),\dots,(Z_{\ncal},\theta_{\ncal}),(Z,\theta)\}$. Note that the value of $\lambda'$ is a deterministic function of $E$. By the weighted exchangeability of the data $\Dcal'$ and the i.i.d. sampling of the models $\{\theta_i\}^{\ncal}_{i=1}\cup\{\theta\}$, we have 
    \begin{align}
        L_{\ncal+1}(\lambda')|E\sim \frac{1}{\sum_{i=1}^{\ncal+1} W_i} \left( \sum_{i=1}^{\ncal+1} W_i \delta_{L_i(\lambda')} \right)
        \label{eq:l_te_distr}
    \end{align}
    where $\delta_{x}$ denotes the Dirac delta function centered at value $x$. From \eqref{eq:l_te_distr}, we thus conclude that 
    \begin{equation}
        \!\!\mathbb{E}\!\left[L_{\ncal+1}( \lambda')|E\right] \!=\! \frac{1}{\sum_{i=1}^{\ncal+1} W_i} \!\left( \sum_{i=1}^{\ncal+1} \!\!W_i L_i(\lambda') \!\right) \!\leq \alpha.\!
    \end{equation}
    Combined with \eqref{eq:risk_ineq_aug} and taking the expectation over the realization of the multiset $E$, we obtain the final result.
    
    \subsection{On the Base Predictor Generalization Error}
    \label{app:gen_base_pred}
    Let $P(\theta)$ be the marginal distribution on $\Theta$ induced by the the product distribution $Q\times P^{\otimes \ntr}_{XY}$. The mutual information between the model's parameter and the training data $\Dtr$ can be expressed as, with the expectation taken over $\Dtr$,
    \begin{align}
        I(\theta;\Dtr)=\mathbb{E}[{\rm KL}(Q||P)]
    \end{align}
    where ${\rm KL}(Q||P)$ is the Kullback-Leibler (KL) divergence between $Q(\theta|\Dtr)$ and $P(\theta)$.
    For many popular learning algorithms such as Gibbs posteriors, $\epsilon$-differentially private algorithms with $\epsilon=O(\log(\ntr)/\ntr)$, and stochastic gradient Langevin dynamics (SGLD), the mutual information scales as \cite{pensia-18a,feldman-18a,raginsky-21a}
    \begin{align}
        I(\theta;\Dtr)=O(\log(\ntr)).
    \end{align}
     Markov's inequality provides the following probabilistic bound on the KL divergence, as per \cite[Lemma 1]{matteo2024informativeness}.
    \begin{lemma}
        \label{lemma:kl}
        Assume that $Q(\theta|\Dtr)$ is absolutely continuous with respect to $P(\theta)$ and that $I(\theta;\Dtr)\leq c\log(\ntr)$ for some $c>0$. Then, for any $\delta\in (0,1)$, with probability at least $1-\delta$, we have the following inequality:
        \begin{align}
            {\rm KL}(Q||P)\leq \frac{I(\theta;\Dtr)}{\delta}\leq  \frac{c\log(\ntr)}{\delta}.
            \label{eq:lemma-kl}
        \end{align}
    \end{lemma}
    We now show that the conditions stated in Lemma \ref{lemma:kl} suffice to satisfy Assumption \ref{ass:gen_base}, thereby guaranteeing that the generalization error bound holds for many popular learning algorithms. 

    Denote the population risk under the training distribution of the set predictor with parameter $\lambda$ as
    \begin{align}\tilde{\mathcal{L}}_Q(\lambda)=\mathbb{E}\left[\mathcal{L}(\Gamma_{\theta}(\tilde X,\lambda),\tilde Y) \given \Dtr\right],
    \end{align}
    where $(\tilde X,\tilde Y)\sim P_{XY}$ and $\theta\sim Q(\theta|\Dtr)$.
    Based on this definition, we decompose the generalization metric as
	\begin{align}
    	\label{eq:decomposition}
		\Delta(Q|\Dtr )\!&=\sup_{\lambda}|\hat{\mathcal{L}}_Q(\lambda|\Dtr) \!-\! \mathcal{L}_Q(\lambda)|\\
        &= \sup_{\lambda}|\hat{\mathcal{L}}_Q(\lambda|\Dtr ) \!-\! \tilde{\mathcal{L}}_Q(\lambda)+\tilde{\mathcal{L}}_Q(\lambda) \!-\! \mathcal{L}_Q(\lambda)|\nonumber\\
        &\leq \underbrace{\sup_{\lambda}|\hat{\mathcal{L}}_Q(\lambda|\Dtr ) \!-\! \tilde{\mathcal{L}}_Q(\lambda)|}_{:=\tilde{\Delta}(Q|\Dtr )}\!+\! \underbrace{\sup_{\lambda}|\tilde{\mathcal{L}}_Q(\lambda) \!-\! \mathcal{L}_Q(\lambda)|}_{:=\Delta_b(Q|\Dtr )}.\!\nonumber
	\end{align}
    The first term $\tilde{\Delta}(Q|\Dtr)$ corresponds to the difference between the empirical risk and population risk under the training distribution $P_{XY}$ while the second $\Delta_b(Q|\Dtr )$ is a bias term corresponding to the difference between the population risk under the training distribution $P_{XY}$ and test distribution $P'_{XY}$.
    \subsubsection{Bounding $\tilde{\Delta}(Q|\Dtr)$}
	Here we focus on the first term of the decomposition \eqref{eq:decomposition}, i.e., the absolute difference over $\lambda$ between the empirical risk and population risk under $P_{XY}$. We use the shorthand notation 
    \begin{align}
        \mathcal{L}_\theta(\lambda,Z)=\mathcal{L}(\Gamma_\theta(X,\lambda),Y).
    \end{align}
    For a fixed model $\theta'\in \Theta$, we denote
	\begin{align}
		\tilde{\Delta}(\theta'|\Dtr )=	\tilde{\Delta}(Q|\Dtr)|\Big|_{Q=\delta_{\theta'}},
	\end{align}
    where $\delta_{\theta'}$ is Dirac delta function centered at model $\theta'$.
    Similarly, let $\hat{\mathcal{L}}_{\theta'}(\lambda|\Dtr )=\hat{\mathcal{L}}_{\delta_{\theta'}}(\lambda|\Dtr )$ and $\tilde{\mathcal{L}}_{\theta'}(\lambda)=\tilde{\mathcal{L}}_{\delta_{\theta'}}(\lambda)$.
	From the Donsker-Varadhan identity, it follows that
	\begin{equation}
     \label{eq:dv_identity}
      \mathbb{E}\left[e^{\mathbb{E}\left[\tilde{\Delta}(\theta|\Dtr )]\given \Dtr \right] - {\rm KL}(Q||P) }\right] \leq	\mathbb{E}\left[e^{\tilde{\Delta}(\tilde \theta|\Dtr )]}\right]
	\end{equation}
        where $\tilde\theta\sim P$.
	For any fixed $\theta'$, using a symmetrization argument with ghost samples $\Dtr'$ drawn from $P_{XY}^{\otimes \ntr}$, 
	\begin{align}
		\mathbb{E}\hspace{-0.25em}\left[e^{\tilde{\Delta}(\theta'|\Dtr )]}\right]\!
        &=
        \mathbb{E}\hspace{-0.25em}\left[e^{\sup_\lambda |\hat{\mathcal{L}}_{\theta'}(\lambda|\Dtr )-\tilde{\mathcal{L}}_{\theta'}(\lambda)| }\right]\nonumber\\
		&=\mathbb{E}\hspace{-0.25em}\left[e^{\sup_\lambda |\hat{\mathcal{L}}_{\theta'}(\lambda|\Dtr )-\mathbb{E}[\hat{\mathcal{L}}_{\theta'}(\lambda|\mathcal{D}'_{\rm tr})]| }\right]\nonumber\\
		&\leq\mathbb{E}\hspace{-0.25em}\left[e^{\mathbb{E}\left[\sup_\lambda |\hat{\mathcal{L}}_{\theta'}(\lambda|\Dtr )-\hat{\mathcal{L}}_{\theta'}(\lambda|\mathcal{D}'_{\rm tr})|\given \Dtr \right]}\right]\nonumber\\
		&\leq\mathbb{E}\hspace{-0.25em}\left[e^{\sup_\lambda |\hat{\mathcal{L}}_{\theta'}(\lambda|\Dtr )-\hat{\mathcal{L}}_{\theta'}(\lambda|\mathcal{D}'_{\rm tr})|}\right]\nonumber\\
		&\leq \mathbb{E}\hspace{-0.25em}\left[e^{2\mathbb{E}\left[ \sup_\lambda \left|\frac{1}{\ntr}\sum^{\ntr}_{i=1}\xi_i\mathcal{L}_{\theta'}(\lambda,Z_i)\right| \given \Dtr \right]}\right] \!
	\end{align}
    where $\{\xi_i\}^{\ntr}_{i=1}$ are Rademacher variables.
    
	The last line includes the empirical Rademacher complexity of the function class induced by the parameter $\lambda$, i.e. $\mathcal G_{\theta'} := \{Z \to \mathcal{L}_{\theta'}(\lambda, Z): \lambda \in [0, \Rmax]\}$. 
    The pseudo-dimension of the function class $\mathcal G_{\theta'}$ is bounded by 1 for all $\theta'$ (see \cite[Def.~11.5]{mohri2018foundations}). The pseudo-dimension bound holds because the loss function is assumed to be non-increasing in $\lambda$, effectively making the function class equivalent to thresholding. Note also that the loss function is non-negative and bounded by $\Lmax$. The empirical Rademacher complexity can then be bounded using VC dimension bounds as \cite[Cor.~3.8 and 3.18]{mohri2018foundations}
	\begin{align}
		\! \mathbb{E}\!\left[e^{\!2\mathbb{E} \left[\sup_\lambda |\frac{1}{\ntr}\sum^{\ntr}_{i=1}\xi_i\mathcal{L}_{\theta}(\lambda,Z_i)| \given \Dtr \right]}\right] \!\leq e^{2\Lmax\sqrt{\frac{2\log(e\ntr)}{\ntr}}}.\!
	\end{align}
	Averaging over $\tilde \theta\sim P$ and combining with \eqref{eq:dv_identity}, we obtain
	\begin{align}
        \label{eq:mean_exp_risk}
		\mathbb{E}\left[e^{\frac{1}{\sqrt{\frac{8\log(e\ntr)}{\ntr}}}\mathbb{E}\!\left[\tilde{\Delta}(\theta|\Dtr )\given \Dtr\right]-{\rm KL}(Q||P)}\right]\leq	e^{\Lmax}.
	\end{align}
	Applying the Chernoff bound to the random variable in \eqref{eq:mean_exp_risk}, noting that $\mathbb{E}[\tilde{\Delta}(\theta|\Dtr )\given \Dtr]\geq\tilde{\Delta}(Q|\Dtr )$, yields the probabilistic guarantee
	\begin{align}
		\Pr_{\Dtr }\hspace{-0.25em}\left[ \frac{\tilde{\Delta}(Q|\Dtr )-{\rm KL}(Q||P)}{\sqrt{\frac{8\log(e\ntr)}{\ntr}}}> \log\frac1\delta+\Lmax\right]\hspace{-0.25em}\leq \delta .
	\end{align}
	Rearranging the terms, we have with probability at least $1-\delta$
        \begin{equation}
            \! \tilde{\Delta}(Q|\Dtr ) \leq \sqrt{\frac{8\log(e\ntr)}{\ntr}}\!\left(\log\frac1\delta+\Lmax+{\rm KL}(Q||P)\right).\!
        \end{equation}
	By a union bound argument, we further bound the KL term as per Lemma \ref{lemma:kl} to conclude that with probability at least $1-\delta$
    \begin{equation}
        \label{eq:prob_bound_est}
            \! \tilde{\Delta}(Q|\Dtr ) \leq \sqrt{\frac{8\log(e\ntr)}{\ntr}}\!\left(\log\frac2\delta+\Lmax+\frac{2c\log\ntr}{\delta} \right).\!
        \end{equation}

	This completes the bound on $\tilde{\Delta}(Q|\Dtr )$.

    \subsubsection{Bounding $\Delta_b(Q|\Dtr)$}
    It remains to bound the bias term $\Delta_b(Q|\Dtr )$ in \eqref{eq:decomposition}. 
    Since the loss $\mathcal{L}(\cdot,\cdot)$ is bounded by $\Lmax$, it is $\Lmax/2$-sub-Gaussian.
    Hence, the bias term can be bounded via the KL divergence between the training marginal $P_X$ and test marginal $P_X'$  \cite[Lemma 1]{xu2017information}. Specifically,
    \begin{align}
         \label{eq:bias_term}
         \Delta_b(Q|\Dtr )\leq \Lmax\sqrt{{\rm KL}(P_X||P'_X)}\leq \Lmax\sqrt{\log(\Wmax)} .
    \end{align}
    Combining \eqref{eq:prob_bound_est}  and \eqref{eq:bias_term} we conclude that 
    \begin{align}
		\Pr_{\Dtr }&\Bigg[ \Delta(Q|\Dtr )]>  \Lmax\sqrt{\log(\Wmax)}+\nonumber\\
        &+\sqrt{\frac{8\log(e\ntr)}{\ntr}}\left(\log\frac2\delta+\Lmax+\frac{2c\log\ntr}{\delta}\right)\hspace{-0.25em}\Bigg]\hspace{-0.25em}\leq \delta.
	\end{align}
    With this, we can conclude that the conditions of Lemma~\ref{lemma:kl} suffice to guarantee Assumption~\ref{ass:gen_base}, ensuring that it holds for many popular learning algorithms.
	\subsection{Proof of Theorem \ref{th:gen_bound}}\label{app:main_thm_pf}
    The expected set size of the W-CRC predictor (\ref{eq:crc_set_predictor}) can be expressed as
\begin{align}
	\Lambdawcrc&=\mathbb{E}[|\Gammawcrc(X\vert \Dfull)|\given \Dtr]\\
	&=\mathbb{E}\left[\int_{\mathcal Y} \mathbbm{1}\{R_\theta(X,y)\leq 	\lambdawcrc(X|\Dfull)\}dy\right]. \nonumber
\end{align}

For any fixed $\theta'$, we denote as $\lambdaythp(X)$ the value of the NC score $R_{\theta'}(X,y)$, recalling that $X\sim P'_X$. 
For all $y\in \mathcal{Y}$ such that $\lambdaythp(X)\leq\lambdawcrc(X|\Dfull)$, the augmented weighted empirical calibration loss satisfies $\hat{\mathcal{L}}_Q(\lambdaythp,X|\Dfull)\geq \alpha$.
It follows that 
\begin{align}
	\!\!\Lambdawcrc\!&\leq \int_{\mathcal Y} \Pr\left[	 \hat{\mathcal{L}}_Q(\lambdayth,X|\Dfull)\geq\alpha\right]dy \\
	&=\!\int_{\mathcal Y} \!\!\Pr\left[	\sum^{\ncal}_{i=1} W_i\left(L_i(\lambdayth)-\alpha\right)\geq W(\alpha-\Lmax)\right] \!dy\nonumber\\
	&\leq\! \int_{\mathcal Y} \!\!\Pr\left[	\sum^{\ncal}_{i=1} W_i\left(L_i(\lambdayth)-\alpha\right)\geq\Wmax(\alpha-\Lmax)\right] \!dy \nonumber
\end{align}
Conditioned on the test NC score value $\lambdayth(X)$, the inefficiency can be expressed as
\begin{align}
    \label{eq:bound_size_nc_score_prob}
	&\frac{\Lambdawcrc}{|\mathcal{Y}|}\leq\frac{1}{|\mathcal{Y}|}\int_{0}^{\infty}dr\int_{\mathcal Y} dy \Pr[\lambdayth(X)=r]\cdot \\
	&\hspace{1em}\cdot\Pr\left[\sum^{\ncal}_{i=1} 	\frac{W_i\left(L_i(\lambda_{y})-\alpha\right)}{\ncal}>\frac{\Wmax(\alpha-\Lmax)}{\ncal} \given \lambdayth(X)=r\right] \nonumber
\end{align}
The first factor of \eqref{eq:bound_size_nc_score_prob} corresponds to the size of the NC score at level $r$ \eqref{eq:exp_mf}.
The expectation of the random variable in the second factor of \eqref{eq:bound_size_nc_score_prob} is, recalling that $W_i$ is the likelihood ratio and $L_{\ncal+1}(r)$ is the expected loss on a sample from $P'_{XY}$,
\begin{align}
	\mathbb{E}\hspace{-0.25em}\left[\sum^{\ncal}_{i=1} \frac{W_i(L_i(r)-\alpha)}{\ncal}\right]\hspace{-0.25em}&=\mathbb{E}\hspace{-0.25em}\left[\sum^{\ncal}_{i=1}\frac{L_{\ncal+1}(r)-\alpha}{\ncal}\right]\hspace{-0.25em}\nonumber\\
    &=\mathcal{L}_Q(r)-\alpha.
\end{align}
Given that the loss function $\mathcal{L}_Q(r)$ is non-increasing in $r$, for values of $r \geq \lambda^*$ with 
\begin{align}
	\lambda^*=\inf \left\{r:\mathcal{L}_Q(r) \leq\alpha+\frac{\Wmax(\alpha-\Lmax)}{\ncal}\right\},
\end{align}
it is possible to leverage concentration inequalities to bound the second factor in \eqref{eq:bound_size_nc_score_prob}.
More specifically, based on Hoeffding's inequality, for $r\geq\lambda^*$ it holds \cite[Propo.~2.5]{wainwright-19a}
\begin{align}
	\Pr&\left[\sum^{\ncal}_{i=1} \frac{W_i\left(L_i(r)-\alpha\right)}{\ncal}>\frac{\Wmax(\alpha-\Lmax)}{\ncal} \given \lambdayth(X)=r \right]\leq\nonumber \\
    &\hspace{1em}\exp\left\{-\frac{2 \ncal(\Wmax(\alpha-\Lmax)/\ncal-
		\mathcal{L}_Q(r)+\alpha)^2}{\Lmax^2\Wmax^2}\right\}.
\end{align}
For values of $r<\lambda^*$, we simply bound the probability by 1. We then obtain the intermediate result
\begin{align}
    \label{eq:intermediate}
	\frac{\Lambdawcrc}{|\mathcal{Y}| }&\leq\int^{\infty}_{\lambda^*}  e^{-\frac{2 \ncal(\Wmax(\alpha-\Lmax)/\ncal-
		\mathcal{L}_Q(r)+\alpha)^2}{\Lmax^2\Wmax^2}} \gamma(r)dr\nonumber\\
        &\hspace{2em}+\int^{\lambda^*}_{0} \gamma(r)dr.
\end{align}
The expression \eqref{eq:intermediate} is increasing in the value $\lambda^*$, which is a function of the base predictor $Q(\theta|\Dtr)$. Define
\begin{align}
	\hat{\lambda}=\inf \Bigg\{r:	\hat{\mathcal{L}}_Q&(r|\Dtr)\leq\\
    &\!\!\alpha+\frac{\Wmax(\alpha-\Lmax)}{\ncal}-\frac{\beta(\delta,\ntr)}{\sqrt{\ntr}}-\Lmax\sqrt{\log(\Wmax)}\Bigg\}.\nonumber
\end{align}
From Assumption \ref{ass:gen_base}, with probability at least $1-\delta$ with respect to the draw of the training data $\Dtr$, we have   $\mathcal{L}_Q(r)\leq\hat{\mathcal{L}}_Q(r|\Dtr)+\frac{\beta(\delta,\ntr)}{\sqrt{\ntr}}+\Lmax\sqrt{\log(\Wmax)}$, and therefore $\lambda^*\leq 	\hat{\lambda}$.
We conclude that with probability at least $1-\delta$, the final bound on the W-CRC efficiency is
\begin{align}
	&\frac{\Lambdawcrc}{|\mathcal{Y}|} \leq\int^{\hat{\lambda}}_{0} \gamma(r)dr+\\
		&\hspace{-0.8em}\int^{\infty}_{\hat{\lambda}}	\hspace{-0.77em}\!e^{-2 \frac{\ncal}{\Lmax^2\Wmax^2}\left(\frac{\Wmax(\alpha-\Lmax)}{\ncal}-
		\hat{\mathcal{L}}_Q(r|\Dtr)+\alpha-\frac{\beta(\delta,\ntr)}{\sqrt{\ntr}}-\Lmax\sqrt{\log(\Wmax)}\right)^2} \!\!\!\gamma(r)dr. \nonumber
\end{align}
\end{appendix}
\end{document}